\documentclass[11pt,a4paper]{article}
\usepackage{times,latexsym}
\usepackage{url}
\usepackage[T1]{fontenc}

\usepackage[acceptedWithA]{tacl2021v1}

\usepackage{xspace,mfirstuc,tabulary}

\newif\iftaclinstructions
\taclinstructionsfalse 
\iftaclinstructions
\renewcommand{\confidential}{}
\renewcommand{\anonsubtext}{(No author info supplied here, for consistency with
TACL-submission anonymization requirements)}
\newcommand{\instr}
\fi

\iftaclpubformat 

\else

\fi

\usepackage{times}
\usepackage{latexsym}

\usepackage[T1]{fontenc}

\usepackage[utf8]{inputenc}

\usepackage{microtype}

\usepackage[nolist]{acronym}
\usepackage{amsmath}
\usepackage{graphicx}
\usepackage{tcolorbox}
\usepackage{color, colortbl}
\usepackage{xcolor}

\usepackage{multirow}
\usepackage{booktabs}
\usepackage{amsmath}
\usepackage{graphics}
\usepackage{graphicx}

\usepackage{caption}
\usepackage{subcaption}
\usepackage{tabularx}
\usepackage{xspace}
\usepackage{arydshln}
\usepackage{booktabs}
\usepackage{balance}

\usepackage{amssymb}
\usepackage{pifont}
\newcommand{\cmark}{\ding{51}}%
\newcommand{\xmark}{\ding{55}}%

\makeatletter
\newcommand*{\org@overidelabel}{}
\let\org@overridelabel\@verridelabel
\@ifpackagelater{acronym}{2015/03/21}{
  \renewcommand*{\@verridelabel}[1]{%
    \@bsphack
    \protected@write\@auxout{}{\string\AC@undonewlabel{#1@cref}}%
    \org@overridelabel{#1}%
    \@esphack
  }%
}{
  \renewcommand*{\@verridelabel}[1]{%
    \@bsphack
    \protected@write\@auxout{}{\string\undonewlabel{#1@cref}}%
    \org@overridelabel{#1}%
    \@esphack
  }%
}
\makeatother

\newcommand{\model}{\textsc{DoT5}\xspace}
\newcommand{\modellarge}{\textsc{DoT5}$_{\text{large}}$}
\newcommand{\modelsequential}{T5$_{\text{large}}$-MLM$\rightarrow$Task\xspace}

\newcommand{\spacy}{SpaCy }

\definecolor{Gray}{gray}{0.9}
\definecolor{Almond}{rgb}{0.94, 0.87, 0.8}
\definecolor{Pink}{rgb}{0.907, 0.762, 0.868}
\definecolor{antiquewhite}{rgb}{0.98, 0.92, 0.84}
\definecolor{cambridgeblue}{rgb}{0.64, 0.76, 0.68}
\definecolor{lightblue}{rgb}{0.68, 0.85, 0.9}
\usepackage{cleveref}
\crefformat{section}{\S#2#1#3}
\crefformat{subsection}{\S#2#1#3}
\crefformat{subsubsection}{\S#2#1#3}
\crefrangeformat{section}{\S\S#3#1#4 to~#5#2#6}
\crefmultiformat{section}{\S\S#2#1#3}{ and~#2#1#3}{, #2#1#3}{ and~#2#1#3}
\usepackage{refstyle}
\Crefformat{figure}{#2Fig.~#1#3}
\Crefmultiformat{figure}{Figs.~#2#1#3}{ and~#2#1#3}{, #2#1#3}{ and~#2#1#3}
\Crefformat{table}{#2Table~#1#3}
\Crefmultiformat{table}{Tabs.~#2#1#3}{ and~#2#1#3}{, #2#1#3}{ and~#2#1#3}
\Crefformat{appendix}{Appx.~\S#2#1#3}
\crefformat{algorithm}{Alg.~#2#1#3}

\title{Compositional Zero-Shot Domain Transfer\\ with Text-to-Text Models}

\author{Fangyu Liu$^{1}$\thanks{\ \ Work done at Microsoft Health Futures.}, Qianchu Liu$^{2}$, Shruthi Bannur$^{2}$, Fernando Pérez-García$^{2}$,\\ \textbf{Naoto Usuyama$^{2}$, Sheng Zhang$^{2}$, Tristan Naumann$^{2}$, Aditya Nori$^{2}$, Hoifung Poon$^{2}$,}\\ \textbf{Javier Alvarez-Valle$^{2}$, Ozan Oktay$^{2}$, Stephanie L. Hyland$^{2}$}\\
$^{1}$ University of Cambridge \ \ \ \ \ $^{2}$ Microsoft Health Futures\\\small{\texttt{fl399@cam.ac.uk, \{t-floraliu,stephanie.hyland\}@microsoft.com}}}

\begin{document}
    \begin{acronym}
    \acro{LLM}{large language model}
    \acro{MLM}{masked language modelling}
    \acro{NEM}{named entity matching}
    \acro{NLG}{natural language generation}
    \acro{NLI}{natural language inference}
    \acro{NLU}{natural language understanding}
    \acro{NLP}{natural language processing}
    \acro{SOTA}{state-of-the-art}
    \acro{STS}{sentence retrieval similarity}
    \acro{T5}{Text-to-Text Transfer Transformer}
    \acro{NLP}{natural language processing}
    \acro{PMC}{PubMed Central}
\end{acronym}

    \maketitle

\begin{abstract}


Label scarcity is a bottleneck for improving task performance in specialised domains.
We propose a novel compositional transfer learning framework (\model%
\footnote{\model (read as ``dot five''): \textbf{D}omain Compositional Zer\textbf{O}-shot \textbf{T5}.}) for zero-shot domain transfer. Without access to in-domain labels, \model jointly learns domain knowledge (from \acl{MLM} of unlabelled in-domain free text) and  task knowledge (from task training on more readily available general-domain data) in a multi-task manner.
To improve the transferability of task training, we design a strategy named NLGU: we simultaneously train \ac{NLG} for in-domain label-to-data generation which enables data augmentation for self-finetuning and \ac{NLU} for label prediction.
We evaluate \model on the biomedical domain and the resource-lean subdomain of radiology, focusing on \acl{NLI}, text summarisation and embedding learning.
\model demonstrates the effectiveness of compositional transfer learning through multi-task learning. In particular, \model outperforms the current \acl{SOTA} in zero-shot transfer by over 7 absolute points in accuracy on RadNLI.
We validate \model with ablations and a case study demonstrating its ability to solve challenging NLI examples requiring in-domain expertise.
\end{abstract}

\acresetall
    \section{Introduction}\label{sec:intro}

While pretrained language models demonstrate massive improvements on a wide range of \ac{NLP} tasks, it remains challenging to apply them to specialised domains \citep{ramponi-plank-2020-neural}.
To acquire domain-specific task knowledge, a conventional approach is to perform domain-specific pretraining --- usually \ac{MLM} on in-domain raw text --- followed by finetuning with in-domain task-annotated data \citep{lee2020biobert,gu2021domain,boecking2022making}.
However, this approach requires in-domain task labels that can be expensive to acquire. Another approach is to train a model with the usually abundant general-domain task labels and directly transfer to the new domain \citep{romanov2018lessons, ma-etal-2021-zero}, but the transfer performance is often limited by the domain gap. Past studies on zero-shot domain transfer or unsupervised domain adaptation have explored methods to transfer task knowledge from a source domain to an unseen target domain \citep{ramponi-plank-2020-neural,ganin2015unsupervised}, but they usually require external modules to perform feature or domain alignment and are not always easily applicable to pretrained language models. In particular, there is little understanding of how we can leverage and combine domain-specific knowledge and general-domain task knowledge in the context of the recent success of text-to-text architectures in transfer learning. 

To close this gap, we propose \model, a novel compositional zero-shot domain-transfer framework based on the \ac{SOTA} transfer learning model \ac{T5} \citep{raffel2019exploring}. Throughout, the `zero-shot' setup refers to zero-shot \emph{domain} transfer with no access to labelled \emph{in-domain} data.\footnote{The definition of `zero-shot' in this paper follows recent studies \citep{pan-etal-2022-task, zhao-etal-2022-domain}, and is similar to unsupervised domain adaptation, as discussed in  \Cref{sec:rw}. Another similar usage of ‘zero-shot' is found in cross-lingual setups where no task labels are accessible in the target test language but labels in the same task are available in a source language. Note that this definition is different from ‘zero-shot learning’ traditionally used to refer to the prediction of unseen classes.} By ``compositional'' we mean that \model{} is able to combine seen task labels and domain text to acquire an unseen combination of task domain knowledge.

\begin{figure}
    \centering
    \includegraphics[width=1\linewidth]{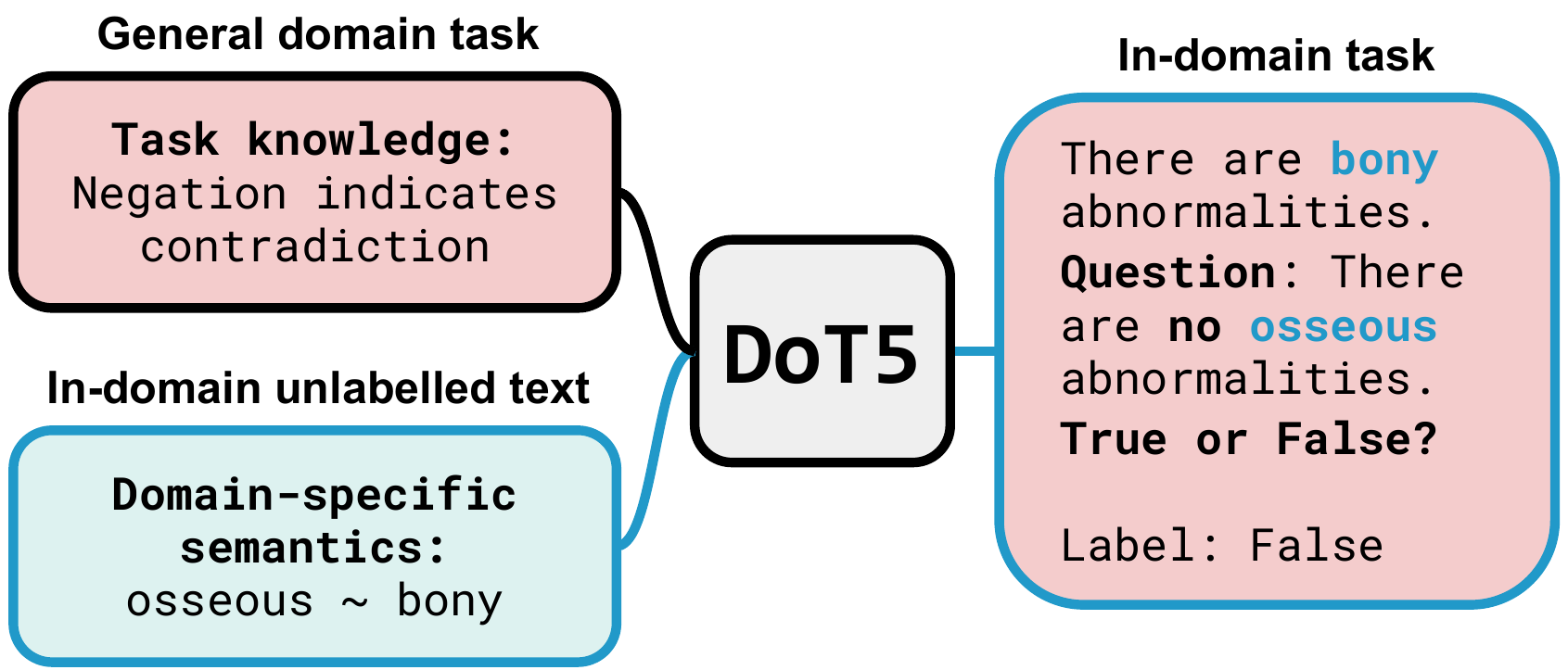}
    \caption{By combining task knowledge from general domain data and domain knowledge from in-domain unlabelled text, our text-to-text model \model learns to solve in-domain tasks.
    }
    \label{fig:front_page}
\end{figure}

As shown in \Cref{fig:front_page},
\model combines domain knowledge and task knowledge by making the best use of in-domain free text and general-domain task labels, which are typically accessible and abundant.
For example, in the context of \ac{NLI}, \model can learn domain-specific semantics (e.g., ``bony abnormalities'' is a synonym of ``osseous abnormalities'') from in-domain free text and transferable task knowledge from general-domain task labels (e.g., negation indicates contradiction) to infer domain-specific task knowledge (e.g., ``There are no bony abnormalities'' contradicts ``There are osseous abnormalities'').

We apply \model to \ac{NLI}, summarisation and text embedding learning, which are fundamental applications across many domains, and we explore zero-shot domain transfer to the high-value and highly-specialised domain of biomedicine and its extremely low-resource subdomain of radiology. Due to their specialisation, obtaining labelled data in these domains is expensive and time-consuming. For example, the radiology-specific \ac{NLI} dataset (RadNLI) \citep{miura-etal-2021-improving} contains only 960 manually-labelled examples as development and test data and no training data is available.

\begin{figure*}[t]
    \centering
    \includegraphics[width=1\linewidth]{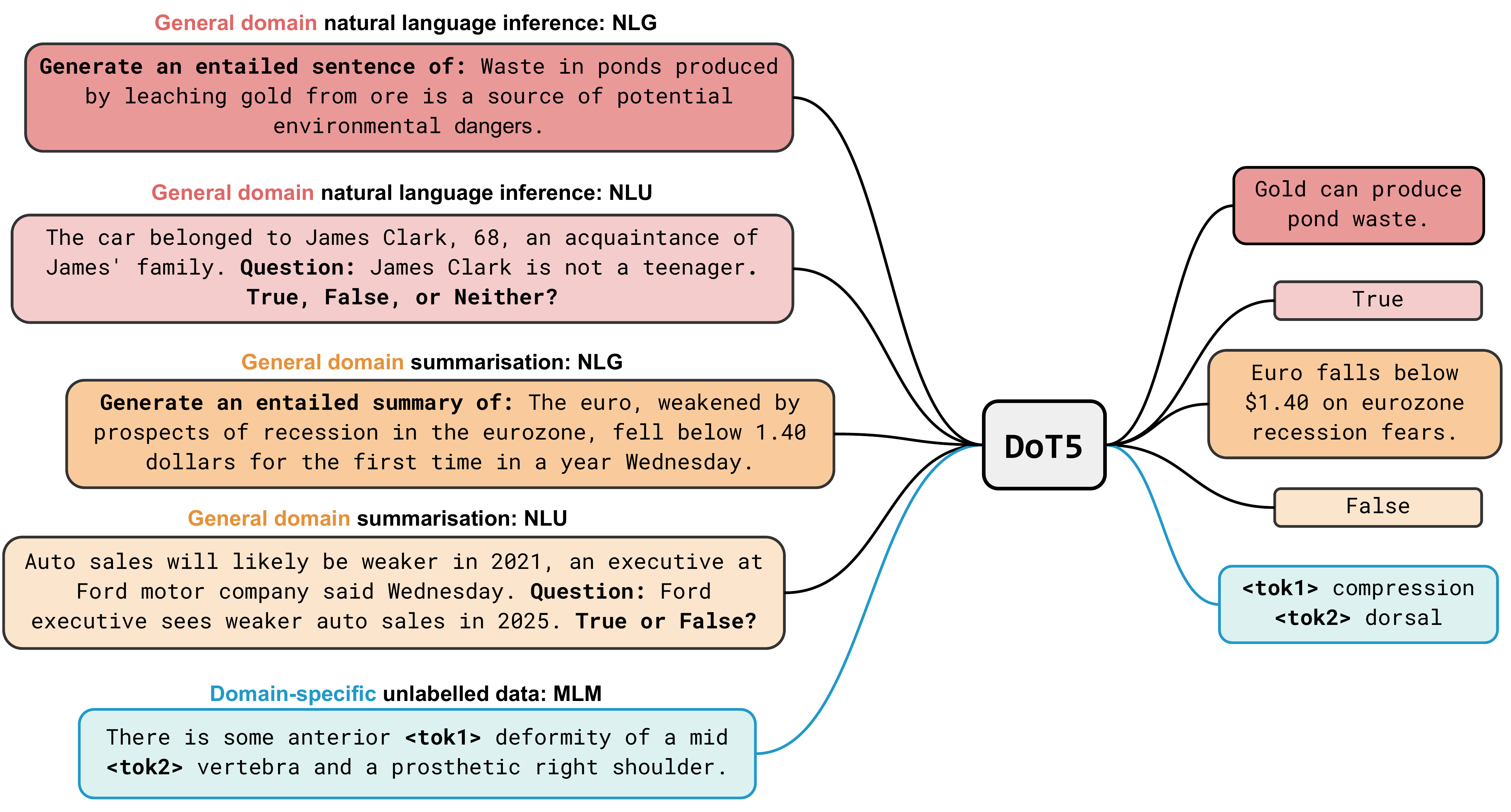}
    \caption{
        Continual pretraining of \model on general-domain tasks (warm colors) and in-domain unlabelled text (blue). For task training, we form both \ac{NLG} and \ac{NLU} variants of \ac{NLI} and summarisation. All training is performed simultaneously, exploiting the unified text-to-text framework of \ac{T5}.
    }
    \label{fig:nli}
\end{figure*}

The key to \model's compositional transfer is {\it continual multi-task pretraining to simultaneously acquire domain and task knowledge}: we jointly train \ac{T5} with \ac{MLM} on in-domain unlabelled data and general-domain tasks (\ac{NLI} and summarisation).
To better acquire the transferable task knowledge from the general-domain task labels, we propose a multi-task setup we call NLGU.
As depicted in \Cref{fig:nli}, NLGU gives each task two formulations:  \ac{NLG} (label-to-data generation), and \ac{NLU} (data-to-label prediction).
\ac{NLU} enables label prediction when tested in an unseen domain and forces model sensitivity to the conditioned label, assisting \ac{NLG}.
Meanwhile, \ac{NLG} enables downstream tasks such as summarisation or data augmentation.
This enables \model to generate its own \ac{NLI} in-domain task data for further finetuning (a process we call self-finetuning), or to generate positive and negative examples for improving text embeddings by contrastive learning \citep{oord2018representation}.

Our experiments show the effectiveness of \model in zero-shot domain transfer, and our proposed multi-task compositional approach achieves large gains compared with sequential training with T5 across all tasks. 
In particular, we achieve \ac{SOTA} zero-shot domain transfer performance on RadNLI~\cite{romanov2018lessons}, outperforming baselines including \acp{LLM}, sequential training approaches and task-specific baselines by large margins. 
We also identify several key insights through extensive analysis:
1) All three key components (in-domain MLM, NLGU, self-finetuning) in \model are important for transfer success while multi-task learning with in-domain MLM is the key for combining domain and task knowledge.
2) Scaling up model size significantly improves transfer performance.
3) \model is able to solve challenging domain-specific task examples, indicating it acquires domain-specific task knowledge through compositional transfer.
 
To summarise, we present the following major contributions:
1) We propose \model, a general framework for compositional transfer learning with text-to-text models, and show multi-task training is superior to sequential training in the models' domain transfer. 
2) With a novel NLGU training strategy combining generation and understanding, \model can be used for both classification and generation tasks.\footnote{Notice that the tasks are limited to those that can have pairwise input instead of single sentence input.}
With the latter, \model can perform self-finetuning to further improve transfer performance.
3) We show the effectiveness of \model in zero-shot domain transfer, achieving \ac{SOTA} zero-shot performance in radiology \ac{NLI}.
4) Comprehensive analysis demonstrates the inner workings of \model's compositional transfer. 

\begin{table*}[t]
\small
\centering
\scalebox{0.83}{
    \centering
    \begin{tabular}{l l l c}
    \toprule
        & Setting & Prompt (Input) & Output \\
        \midrule
        \multirow{2}{*}{\rotatebox{90}{\ac{NLI}}} & \ac{NLG}: $(x_1, y) \rightarrow x_2$ &   \textbf{Generate a} \{\texttt{label}\} \textbf{sentence of:} \{\texttt{premise}\} & \{\texttt{hypothesis}\} \\
        & \ac{NLU}: $(x_1, x_2)\rightarrow y$ &
        \{\texttt{premise}\} \textbf{Question:} \{\texttt{hypothesis}\} \textbf{True, False or Neither?} & \{\texttt{True | False | Neither}\} \\
        \midrule
        \multirow{2}{*}{\rotatebox{90}{Sum.}} & \ac{NLG}: $(x_1, y) \rightarrow x_2$ &  \textbf{Generate a} \{\texttt{label}\} \textbf{summary of:} \{\texttt{document}\} & \{\texttt{summary}\} \\
        & \ac{NLU}: $(x_1, x_2) \rightarrow y$  &     \{\texttt{document}\} \textbf{Question:} \{\texttt{summary}\} \textbf{True or False?} & \{\texttt{True  | False}\}\\
        \bottomrule
        
    \end{tabular}
    }
    \caption{
        Prompts used for task-specific training with NLGU for both \ac{NLI} and summarisation (Sum).
        For \ac{NLI} $x_1$: premise, $x_2$:  hypothesis, and the label ($y$) is one of \{entailed, neutral, contradictory\}.
        For summarisation $x_1$: document, $x_2$: summary, and the label ($y$) is one of \{entailed, contradictory\}.
    }
    \label{tab:prompts_and_outputs}
\end{table*}
    \section{Related Work}
\label{sec:rw}

\paragraph{Cross-task transfer with text-to-text models}
\ac{T5} \citep{raffel2019exploring} unifies \ac{NLP} tasks under a seq-to-seq framework and solves them using a single model.
T0 \citep{sanh2022multitask}, FLAN \citep{wei2022finetuned}, MetaICL \citep{min2021metaicl}, and ExT5 \citep{aribandi2022ext} build on top of this idea and explore pretraining \ac{T5} with a massive collection of \ac{NLP} datasets with diverse natural language prompts.
Among them, T0, FLAN, and MetaICL investigate pretraining on a set of tasks, and then zero-shot transfer to another set of unseen tasks.

\paragraph{Domain-specific pretraining}
\citet{gururangan-etal-2020-dont} shows continual training on domain and task data can adapt pretrained models for new domains and tasks. Both BioBERT \citep{lee2020biobert} and BlueBERT \citep{peng-etal-2019-transfer} apply the BERT pretraining protocol (i.e., masked language modelling and next sentence prediction) on \ac{PMC} or PubMed articles.
They continue pretraining BERT checkpoints instead of training from scratch. \citet{gu2021domain} demonstrates the importance of domain-specific vocabulary and pretraining from scratch when in-domain text is abundant, and produces PubMedBERT by pretraining on PubMed articles.
Similar to PubMedBERT, SciBERT \citep{beltagy-etal-2019-scibert} pretrains from scratch on a mix of both PMC and computer science publications. \citet{boecking2022making} introduces CXR-BERT which is pretrained on biomedical and radiology corpora. 
SciFive \citep{phan2021scifive} continually pretrains T5 checkpoints on PubMed abstracts with seq-to-seq MLM. 
We compare to finetuned versions of SciFive, PubMedBERT and CXR-BERT in \Cref{sec:main_results}.

\paragraph{Zero-shot domain transfer learning}
Training in one domain and directly testing on another domain has been a prevalent paradigm in zero-shot cross-domain transfer \citep{miura-etal-2021-improving,boecking2022making,agrawal2022large}.
A similar zero-shot setup is also frequently seen in other transfer learning scenarios such as cross-lingual zero-shot learning \citep{conneau-etal-2018-xnli, conneau-etal-2020-unsupervised}. Our summarisation experiment is most similar to such a direct zero-shot setup. 
Concurrently, 
\citet{pan-etal-2022-task} also proposes to combine in-domain training and out-of-domain task knowledge. They proposed a zero-shot in-domain question answering model by finetuning a general-domain RoBERTa model with first domain-specific NER and then general-domain question answering. 
This study is the closest to our approach, with several key differences: Their method requires in-domain labels (in-domain NER) whereas we do not require any in-domain task labels. They only test on question answering whereas we show a more diverse range of evaluation datasets. Additionally, they do sequential training whereas we perform multi-task training. Finally, their model is not generative and therefore it cannot perform NLGU and self-finetuning as we did in our approach (see \Cref{sec:method}).

Our proposed NLGU and self-finetuning strategies are closely related with cross-domain data augmentation. A line of work in information retrieval generates ``in-domain'' pseudo training data leveraging unlabelled in-domain texts.
As an example, \citet{ma-etal-2021-zero,wang-etal-2022-gpl} train a passage-to-query generator for synthesising in-domain queries for the task of zero-shot passage retrieval. Similarly, The NLG component in our proposed NLGU strategy can also perform data augmentation but with better granularity and diversity as we can generate label-conditioned task data to create both positive and negative examples.

Besides zero-shot transfer in NLP, unsupervised domain adaptation (which also assumes labels in current domain and unlabelled data in the target domain) is a long-standing research topic in machine learning in general \citep{huang2006correcting,pan2010domain,ganin2015unsupervised, ramponi-plank-2020-neural}. Many conventional unsupervised domain adaptation methods require external components to align domains on the feature/embedding level. For example, \citet{pan2010cross} proposes applying spectral feature
alignment to align domain-specific words across domains into unified clusters. \citet{ganin2015unsupervised} adds a domain classifier that promotes domain-invariant features via a gradient reversal layer. These methods are not always immediately suitable for the recent pretrained language models especially the text-to-text models. In comparison, our approach exploits the task unifying nature of text-to-text models which contain the inherent transfer learning abilities and requires minimal architecture changes.
    \section{Method}
\label{sec:method}

To achieve compositional transfer, \model acquires domain knowledge and task knowledge via continual pretraining (see \Cref{fig:nli}).
Specifically, we optimise a joint loss function composed of an in-domain masked language model loss (``domain-\ac{MLM}'') and a general-domain task-specific loss:

\begin{equation}
    \mathcal{L}_{\text{joint}} = \lambda  \mathcal{L}_{\text{domain-MLM}} + (1 - \lambda) \mathcal{L}_{\text{task}}
    \label{eq:joint}
\end{equation}
We set $\lambda = 0.5$ but explore tuning it in \Cref{sec:setup}.

We use \ac{T5}, an encoder-decoder generative language modelling framework \citep{raffel2019exploring}, to learn a conditional sequence generator $P(\text{output}|\text{input})$.
\ac{T5} is chosen for two reasons:
1) It is a strong transfer learning model, and
2) it can unify classification and generation, which has potential to further boost transfer performance (see NLGU discussion in \Cref{sec:gen_dom_task_pretr}). 
We use the same pretraining objective (cross-entropy with teacher-forcing) as in \ac{T5}.

We detail the two loss components for continual pretraining in \cref{sec:in_dom_unsup_pretr,sec:gen_dom_task_pretr}.
Once the model has been continually pretrained, it can be used to perform zero-shot domain transfer on a task.
Task-specific designs for inference are given in \cref{sec:task_specific_inference}.

\subsection{Continual Pretraining with In-domain \ac{MLM}}
\label{sec:in_dom_unsup_pretr}

For $\mathcal{L}_{\text{domain-MLM}}$ we use the \ac{MLM} loss \citep{devlin2019bert} to continually pretrain a \ac{T5} on in-domain free text:
Given a piece of sampled radiology or biomedical text, we randomly mask 15\% of its tokens and ask the model to denoise the masked input sequence, i.e., generate the masked tokens.

\subsection{Continual Pretraining on General-domain Tasks}
\label{sec:gen_dom_task_pretr}
For $\mathcal{L}_{\text{task}}$, we define ($x_1$, $x_2$) as a text pair that denotes (\textit{premise}, \textit{hypothesis}) for \ac{NLI}, and (\textit{document}, \textit{summary}) for summarisation.
The standard \ac{NLI} task assigns labels from $y$: \{entailment, neutral, contradiction\}, and the task is $(x_1, x_2) \rightarrow y$.
For summarisation, the task is usually cast as $x_1 \rightarrow x_2$.
We follow \citet{sanh2022multitask} to adopt a multi-task learning strategy to train summarisation and \ac{NLI} simultaneously.
Hence, the basic setup of task learning would be: \ac{NLI} as a discriminative \ac{NLU} task plus summarisation as an \ac{NLG} task. 

\paragraph{NLGU: Simultaneous \ac{NLG} and \ac{NLU}}
One immediate question is whether we can turn each task into both \ac{NLG} and \ac{NLU} (i.e., adding \ac{NLG} for \ac{NLI} and \ac{NLU} for summarisation).
For \ac{NLI}, we can add label-to-data \ac{NLG} to generate pseudo in-domain text for data augmentation, performing $(x_1,y)\rightarrow x_2$ (the label $y$ is used as control code).
For summarisation, we can also follow \ac{NLI} to add a \ac{NLU} task that predicts whether a document-summary pair is entailed (the correct match) or contradictory (a counterfactual summary) (\Cref{sec:setup}).
This \ac{NLU} component aims to improve the factuality of generated text as it encourages the model to distinguish counterfactuals and true summaries. With the hypothesis that performing \ac{NLG} and \ac{NLU} simultaneously will mutually benefit each other, we propose NLGU, meaning joint training of \ac{NLG} and \ac{NLU}.
With NLGU, we unify both summarisation and \ac{NLI} into $(x_1, x_2)\rightarrow y$ for \ac{NLU} and $(x_1,y)\rightarrow x_2$ for \ac{NLG}.
The conditional generator then simultaneously optimises two losses:
\begin{equation}
     \mathcal{L}_{\text{task}} = \gamma \mathcal{L}_{(x_1, x_2)\rightarrow y} + \mathcal{L}_{(x_1,y)\rightarrow x_2}  
\end{equation}
We set $\gamma=10$ to balance the two losses since $x_2$ is usually much longer than $y$ (the classification label).
\ac{NLU} and \ac{NLG} are both trained with sequence-to-sequence generation, and differ only in the input prompt and the expected output (\Cref{tab:prompts_and_outputs}).
The prompt for $\mathcal{L}_{(x_1, x_2)\rightarrow y}$ is from \citet{brown2020language}.
The prompts for summarisation are akin to those for \ac{NLI}, with \texttt{premise} and \texttt{hypothesis} replaced with \texttt{document} and \texttt{summary} respectively, and we only use \{entailment, contradiction\} relations.

\subsection{Task-specific Designs for In-domain Zero-shot Inference}
\label{sec:task_specific_inference}

After continual pretraining, we zero-shot-transfer the trained model to three applications in specialised domains without requiring labels from these domains:
1) \ac{NLI},
2) summarisation, and
3) text embedding learning.

\paragraph{\ac{NLI} (with self-finetuning)}
While the model is capable of directly performing \ac{NLI} after training on general-domain \ac{NLI} task labels with $(x_1, x_2)\rightarrow y$,  we propose an additional step, self-finetuning, to boost transfer performance (\Cref{sec:ablation}).
We first use the model's \ac{NLG} capabilities to generate pseudo in-domain \ac{NLI} data:
We sample a set of sentences from the target domain as premises, and prompt the pretrained model to generate hypotheses (the \ac{NLG} task) with each of the three control codes (labels).
This pseudo-in-domain \ac{NLI} dataset is then used as additional training data to finetune the same model to perform the NLU task: $(x_1,x_2) \rightarrow y$.
The resulting finetuned model is then used for zero-shot \ac{NLI} transfer.

\paragraph{Text summarisation}
We directly prompt the model after continual pretraining to summarise in-domain documents.
We use the same prompt as pretraining:
``Generate an entailed summary of: \{\texttt{document}\}''.
The output summary is then compared against the gold summary.
Since this is already a task of text generation, i.e., $(x_1, y) \rightarrow x_2$, we cannot exploit self-finetuning as for \ac{NLI} since we cannot improve generation from training on the model's own generated pseudo data. 

\paragraph{Text embedding learning}

\model can be directly used as a generator for data augmentation. Apart from creating more pseudo NLI task data to improve NLI, \model can improve domain-specific embedding learning in general. To do so, we sample a set of in-domain sentences as anchors, and prompt the trained model to generate entailed and contradictory sentences to form positive and negative pairs for each anchor.
With beam search size of five, we sample the top-$k$ most probable sequences as the entailed (positives) and contradictory (negatives) sentences of the anchor%
\footnote{We experimented generating one, three, and five pairs of positives and negatives and found three is the best in our setup. We thus use three across all models.
}. Given the collected anchors and positive/negative sentences, we finetune a \ac{SOTA} sentence embedding model with a contrastive loss.
Specifically, we continually finetune the all-mpnet-base-v2\footnote{\scriptsize{\url{https://discuss.huggingface.co/t/train-the-best-sentence-embedding-model-ever-\\with-1b-training-pairs/7354}}. \\ \scriptsize\url{https://huggingface.co/sentence-transformers/all-mpnet-base-v2}
} model with a variant of InfoNCE~\citep{oord2018representation} modified to handle multiple positives~\citep{miech2020end}. The learned embedding space is then used for query-document retrieval or for computing text similarity.

    \section{Experiment}
\label{sec:exp}

We introduce our experimental setup in \Cref{sec:setup}, briefly discuss baseline approaches in \Cref{sec:baselines} and then present results in \Cref{sec:main_results}.

\subsection{Experimental Setup}
\label{sec:setup}

Details of the datasets used for training and evaluation are given in \Cref{tab:datasets}.

\paragraph{Pretraining datasets}

\newcommand{\rotbf}[1]{\rotatebox{90}{\textbf{#1}}}
\newcommand{\raiseh}[1]{\raisebox{#1\height}}

\begin{table}
\small
\setlength{\tabcolsep}{4pt}
\centering
\scalebox{0.75}{
    \begin{tabular}{@{}cllll@{}}
    \toprule
    & \textbf{Dataset} & \textbf{Task} & \textbf{Used for} &
    \textbf{\# Examples}\\
    \midrule
    \multirow{4}{*}{\raiseh{-0.5}{\rotbf{General}}} & SNLI {\scriptsize -- \citeauthor{bowman-etal-2015-large}}  & \ac{NLI} & Task pretrain. & 550K \\
    & MultiNLI {\scriptsize -- \citeauthor{williams-etal-2018-broad}} & \ac{NLI} & Task pretrain. & 392K \\
    & AdversarialNLI {\scriptsize -- \citeauthor{nie-etal-2020-adversarial}}  & \ac{NLI} & Task pretrain. & 162K \\
    & Gigaword {\scriptsize -- \citeauthor{graff2003english}} &  Summ. & Task pretrain. & 1M \\
    
    \addlinespace[2.5pt]
    \multirow{4}{*}{\raiseh{-1.0}{\rotbf{Radiology}}} &
    \multirow{2}{12em}{MIMIC-CXR {\scriptsize -- \citeauthor{johnson2019mimic}}} &  \multirow{2}{4em}{MLM} & \multirow{2}{4em}{Domain pretrain.} & \multirow{2}{4em}{227K}\\
    \\
    & RadNLI {\scriptsize -- \citeauthor{miura-etal-2021-improving}} &  \ac{NLI} & Evaluation & 480 \\
    & Open-I {\scriptsize -- \citeauthor{demner2016preparing}} & Summ. & Evaluation & 683 \\
    
    \addlinespace[2.5pt]
    \multirow{5}{*}{\raiseh{-1.0}{\rotbf{Biomedical}}}  &
    \multirow{2}{8em}{PubMed Abstracts}  & \multirow{2}{4em}{MLM} & \multirow{2}{4em}{Domain pretrain.} & \multirow{2}{4em}{4.2M}\\
    \\
    & MedNLI {\scriptsize -- \citeauthor{romanov2018lessons}} &\ac{NLI} & Evaluation &  1.4K\\
    & PubMed `ShortSum'  & Summ. & Evaluation & 5K\\
    & MedSTS {\scriptsize -- \citeauthor{yanshan2020medsts}} & Similarity & Evaluation & 371\\
    \bottomrule
    \end{tabular}
    }
    \caption{
        Datasets used in the study for task/domain pretraining (`pretrain.') and evaluation.
        The tasks are \ac{NLI}, text summarisation (`Summ.'), and document retrieval based on text embedding similarity. 
    }
    \label{tab:datasets}
\end{table}

\begin{figure}
    \centering
    \includegraphics[width=1\linewidth]{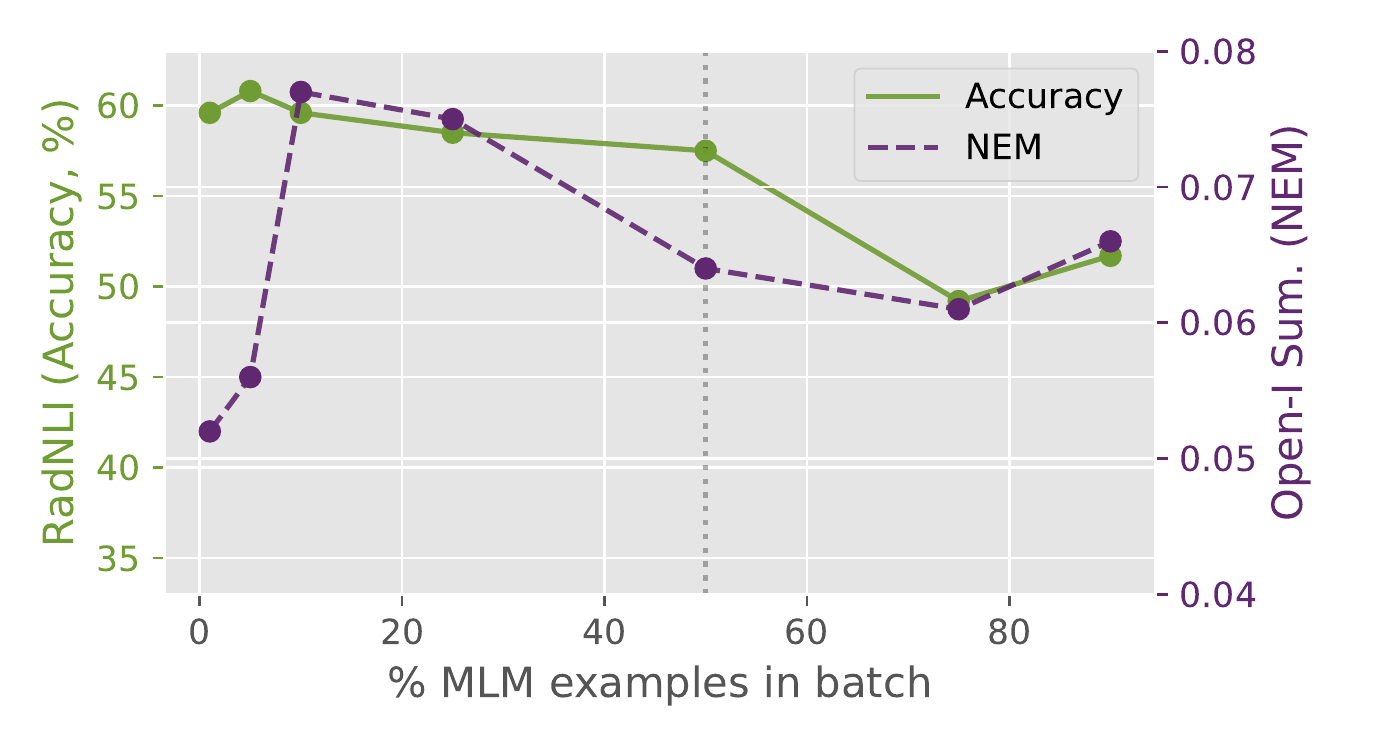}
    \caption{Varying the prevalence of in-domain \ac{MLM} and task data in \model$_\text{small}$ training.}
    \label{fig:joint_loss_weighting}
\end{figure}

As our continual pretraining is a multi-task process, we balance the in-domain and general-domain datasets in each batch via up/downsampling as needed:
For radiology, we upsample MIMIC-CXR samples via duplication, whereas for biomedicine we downsample PubMed abstracts, in each case matching the general-domain task dataset size.
We also balance the number of samples coming from each task, downsampling the summarisation dataset to roughly match that of NLI (`\# Examples' in \Cref{tab:datasets}). Experiments with \model$_\text{small}$ (\Cref{fig:joint_loss_weighting}) indicate that downstream task performance could be boosted by tuning the relative prevalence of the data sources, with a task-dependent optimal value. In this proof-of-concept study, we fix a ratio of 1:1.

We generate counterfactual summaries of Gigaword based on \citet{rajagopal2022counterfactual}. Specifically, we run a named entity recognition model on the documents from the Gigaword summarisation training data, specifically the ``\texttt{en\_core\_web\_sm}'' trained \spacy pipeline~\cite{Honnibal_spaCy_Industrial-strength_Natural_2020}. For each document that contains a named entity, we randomly sample an entity and replace it with a different named entity of the same category from the training corpus. This is our `counterfactual' example\footnote{Note that a counterfactual is not always a contradiction. We approximate contradiction this way and use the `contradictory' control code in our experiments for consistency.}. We also filter out noisy data when the generated counterfactual contains UNK or \#. The resulting dataset, as listed in \Cref{tab:datasets}, consists of 50\% document-`wrong summary' pairs (i.e.\ 500k pairs), one for each true document-summary pair. 
To create pseudo NLI data for the self-finetuning process, we use all premises from the RadNLI/MedNLI development set and generate one entailed, one neutral, and one contradictory hypothesis for each premise. In total, we have 1440 and 4185 pseudo examples for RadNLI and MedNLI respectively. 

\paragraph{Evaluation datasets and metrics}
All the evaluation datasets are from domain-specific tasks (\Cref{tab:datasets}).
For NLI, we report accuracy and macro-$F_1$ (out of 100, for legibility) on the test set of RadNLI and MedNLI.
For summarisation in radiology, we evaluate on findings-to-impression%
\footnote{In a radiology report, the "findings" section is a detailed description and the "impression" section is a summary of the findings with follow-up recommendation.
} summarisation on the test split of the Open-I dataset \cite{demner2016preparing}.
For biomedical summarisation, we create an abstract-to-title summarisation dataset, `PubMed ShortSum'.
The data for this task is sampled from PubMed and filtered to abstracts shorter than 1000 characters.
Compared with the traditional article-to-abstract PubMed summarisation task which evaluates long summary generation for long text \cite{cohan-etal-2018-discourse}, PubMed ShortSum evaluates extreme summarisation for short text and is a more comparable task to our general domain Gigaword summarisation.   
For summarisation evaluation, we use standard lexical metrics (BLEU-4, ROUGE-L) and domain-specific factuality metrics: \ac{NEM} for both radiology \cite{miura-etal-2021-improving} and biomedical \cite{alambo2022entity} summarisation, and CheXbert \citep{smit-etal-2020-combining}%
\footnote{The average of the weighted-$F_1$ score across 14 pathological observations labelled by CheXbert.
} for radiology. 

We evaluate embeddings trained for the biomedical domain on MedSTS \citep{yanshan2020medsts}, a clinical text similarity benchmark.
Since the radiology domain has no text similarity datasets available, we design an impression-to-findings retrieval task on the Open-I test set, and report Accuracy@1/5/10.
This retrieval task can also evaluate embedding quality as it requires the model to differentiate text from same/different reports by encoding texts from matching findings-impression pairs (from the same report) with similar representations. 

\paragraph{Training details}
Models are trained for 10 epochs with validation loss used for checkpoint selection. We use distributed data parallelism on eight GPUs with the largest batch size permissible given computational constraints, resulting in batch sizes of 1024, 512, and 128 for small, base, and large models. With a dataset of  $\sim$8M samples, we thus train the large model for $\sim$64,000 steps per epoch.
We use AdaFactor \citep{shazeer2018adafactor} with learning rates of $10^{-3}$ for MIMIC-CXR and $2\times10^{-5}$ for PubMed pretraining.

\subsection{Baselines}\label{sec:baselines}
We have three categories of baselines: 
(1) task-specific zero-shot baseline models reported from the literature (where applicable);
(2) \acp{LLM} including T0 and GPT-3;
(3) sequential training first on in-domain unlabelled data and then on general-domain task labels. 
All the baseline models in our study must satisfy one constraint: not using any in-domain labels for the task, but they may differ in the required training resources (detailed comparison is found in \Cref{tab:baseline_comparison}). 
We compare with (2) as \acp{LLM} are known to be excellent zero-shot and few-shot learners for an unseen task, and should serve as a reasonable baseline for domain transfer.
We provide (3) as a straightforward baseline to \emph{sequentially} combine in-domain MLM training and general-domain task training as opposed to our proposed multi-task training.

\begin{table}
\small
\setlength{\tabcolsep}{4pt}
\centering
\scalebox{0.9}{
    \begin{tabular}{llll@{}}
    \toprule
    \textbf{Baselines} & \textbf{In-domain} & \textbf{General domain} &  \\
     & \textbf{Text} & \textbf{\ac{NLI}/Summ.}  \\
    \midrule
    \rowcolor{cambridgeblue}
     BERT {\scriptsize -- \citeauthor{miura-etal-2021-improving}}  & \cmark & $\sim$ / $-$  \\
    \rowcolor{cambridgeblue}

    CXR-BERT {\scriptsize -- \citeauthor{boecking2022making}} & \cmark & $\sim$ / $-$   \\
    \rowcolor{cambridgeblue}

    ESIM {\scriptsize -- \citeauthor{Chen2017EnhancedLF}}  & \xmark & \cmark / $-$  \\

    \rowcolor{Gray}
    T0 \& T0++ &  \xmark & \xmark / \cmark  \\
    \rowcolor{Gray}
    GPT-3 & \xmark & \xmark / \xmark  \\
    \rowcolor{Gray}
    GPT-3-\{NLI, GW\} & \xmark & \cmark / \cmark   \\

    \rowcolor{Pink}
    CXR-BERT-NLI & \cmark& \cmark / $-$  \\
    \rowcolor{Pink}
    PubMedBERT-NLI & \cmark& \cmark / $-$  \\
    \rowcolor{Pink}
    SciFive$_{\text{large}}$-\{NLI, GW\} & \cmark & \cmark / \cmark  \\
         \rowcolor{Pink}
     \modelsequential & \cmark & \cmark / \cmark  \\ 
     \model & \cmark & \cmark / \cmark \\

    \bottomrule
    \end{tabular}
    }
    \caption{
        Baseline comparisons grouped into three categories: (1) task-specific zero-shot baselines \colorbox{cambridgeblue}{(green)}, (2) large language models \colorbox{Gray}{(grey)} and (3) sequential training on in-domain text and general-domain task labels \colorbox{Pink}{(pink)}. "\cmark" and "\xmark" specify whether the given data source was used for training. "Summ." means summarisation. Models only evaluated on NLI do not require summarisation data, hence "$-$". "$\sim$" indicates that BERT and CXR-BERT were fine-tuned on MedNLI, a `near-domain' NLI dataset. 
    }
    \label{tab:baseline_comparison}
\end{table}
\paragraph{Task-specific zero-shot baselines}
We compare with the strongest task-specific zero-shot models from the literature.
For the \ac{NLI} task, we compare with \citet{miura-etal-2021-improving} and \citet{boecking2022making}, which both finetune a BERT model with MedNLI training data and then test on RadNLI.
\citet{boecking2022making} performs better as they use radiology-specific BERT model.
Note that MedNLI is a \textit{nearby-domain} corpus rather than general-domain task data, and in fact there has not been successful attempts in the literature to transfer general-domain \ac{NLI} to RadNLI. Note that in the later sequential training section we will establish such baselines from finetuning CXR-BERT on general-domain \ac{NLI}. 
For MedNLI, we compare with the best transfer learning results so far, ESIM (MultiNLI) which was trained on MultiNLI datasets \citep{romanov2018lessons}.
For radiology summarisation, to our knowledge, we are the first to report results on direct transfer from general-domain summarisation.
For biomedical summarisation, since we use a new dataset (PubMed ShortSum), there is no prior comparison.

\paragraph{\Aclp{LLM}}
T0~\citep{sanh2022multitask} and {GPT-3}~\citep{brown2020language} are massively pretrained language models that can be used off-the-shelf for zero-shot or few-shot inference.
T0 is pretrained with multiple tasks including general-domain summarisation datasets (but \emph{not} \ac{NLI}), and shows strong transfer ability \citep{sanh2022multitask}. T0 can be seen as a strong general-domain summarisation model and also strong zero-shot domain transfer baseline on summarisation. T0 is also particularly effective in transferring to unseen tasks. Therefore, we include T0 as a zero-shot baseline for NLI even though it has not been trained with any NLI data.
We test T0 (3B) and the most powerful T0++ (11B) model.
GPT-3 \citep{brown2020language} (\texttt{davinci}) is a massive language model with 175B parameters, pretrained on raw text with an autoregressive language modelling objective.

In the general domain, both models are shown to have performed reasonably well on \ac{NLI} and summarisation with prompting.
We test their zero-shot-inference capabilities in our experiments, following the original papers for prompt design. For the \ac{NLI} task, both T0 models and GPT-3 use the ANLI prompt template described in \citet{brown2020language}: "<premise> Question: <hypothesis> True, False or Neither?". For the summarisation task, T0 used the prompt: "<document> $\backslash$n === $\backslash$n Generate a title for this article:". For GPT-3 summarisation, we used the prompt ("<document>$ \backslash$n$ \backslash$n Tl;dr:") as recommended in the OpenAI GPT-3 playground example\footnote{\scriptsize\url{https://beta.openai.com/examples/default-tldr-summary}}.
Since GPT-3 benefits when few-shot examples are incorporated in the prompt, we create two additional baselines (GPT-3-NLI and GPT-3-GW%
\footnote{These are still zero-shot baselines as they do not use in-domain task examples.
}) that perform in-context learning of the task from general-domain \ac{NLI} training data (30 examples, randomly selected) and Gigaword summarisation training data (20 examples, randomly selected) respectively (\Cref{tab:datasets}).

\paragraph{Sequential training \label{par: sequential training}}
The most straightforward way to exploit both in-domain unlabelled data and task labels is to first train on in-domain MLM and then further finetune on general-domain task labels\footnote{This baseline category is similar to contemporaneous work \cite{pan-etal-2022-task} where domain-task transfer is achieved through sequential in-domain \emph{off-task} training followed by general-domain \emph{in-task} training. Here we do not use in-domain task data of any kind.}. We provide two variants of this baseline. The first type performs continual training with general-domain task labels from \ac{SOTA} domain-specific pretrained models. We adopt SciFive~\cite{phan2021scifive}, a T5 model pretrained on large biomedical corpora, CXR-BERT-General~\cite{boecking2022making}, a radiology-specialised BERT model, and the PubMed-specific PubMedBERT~\cite{gu2021domain}. For finetuning these models we use the same general-domain task data as provided to \model, where for the BERT models we only do finetuning on NLI. This results in baseline models SciFive$_{\text{large}}$-NLI, SciFive$_{\text{large}}$-GW (summarisation), CXR-BERT-NLI, and PubMedBERT-NLI. We further improve SciFive$_{\text{large}}$-NLI by including our proposed self-finetuning stage (SciFive$_{\text{large}}$-NLI + SFT). Since there is no radiology-pretrained T5 model, we compare with SciFive on both domains.

The second baseline type strictly compares multi-task training (\model) and sequential training. Here, we first pretrain T5 with in-domain MLM, and then continually pretrain on the general-domain task data, ensuring other factors remain the same including the training duration, use of NLGU, and use of self-finetuning where appropriate. We call this setting \modelsequential.

\subsection{Main Results}
\label{sec:main_results}

\paragraph{\ac{NLI} (\Cref{table:zeroshot_nli})}

\modellarge\ establishes new \ac{SOTA} for zero-shot domain transfer on RadNLI and competitive results on MedNLI (\Cref{table:zeroshot_nli}).
On RadNLI, \modellarge\ reaches an impressive 82.1\% on accuracy and is the best performing model. It outperforms the strongest reported number from the literature (CXR-BERT) by more than 15\%, and our  baseline CXR-BERT-NLI by almost 7\%. Comparing \model to \modelsequential on RadNLI reveals the benefit of \emph{multitask} training for compositional transfer.

On MedNLI, \modellarge\ outperforms ESIM (MultiNLI) by almost 20\% (accuracy), but does not quite reach the 75.7\% accuracy achieved by PubMedBERT-NLI, which establishes a new \ac{SOTA} in zero-shot domain transfer on MedNLI --- supervised \ac{SOTA} is 86.6\% \citep{phan2021scifive}. Although factors such as tokenisation and pretraining strategies may contribute, we speculate that the domain gap between MedNLI and our biomedical pretraining corpus explains the weaker performance of \model on MedNLI. MedNLI was sourced from \emph{clinical} notes in MIMIC-III, which differ distributionally from biomedical articles in PubMed. Supporting this hypothesis, we observed that \model pretrained on radiology text, and the \emph{sequential} baseline \modelsequential achieved similar performance on MedNLI (70\% accuracy), indicating that results on MedNLI may not fully reflect compositional domain knowledge transfer in our setup. In this case, a strong NLI-specific model is most performant, while lacking potentially-advantageous versatile text generation/summarisation capabilities.

\begin{table}[tbh]
    \small
    \centering
    \scalebox{0.86}{
    \begin{tabular}{@{}lcc@{}}
        \toprule
        \textbf{Model} & \textbf{Accuracy} & \textbf{$F_1$-score}  \\
        \midrule
        
        \multicolumn{3}{c}{\textbf{Radiology (RadNLI)}} \\
        \rowcolor{cambridgeblue}
        BERT \cite{miura-etal-2021-improving}  & 53.3 & -\\ 
        \rowcolor{cambridgeblue}
        CXR-BERT \cite{boecking2022making} & 65.2 & -\\
      
        \rowcolor{Gray}
        T0 (3B)  & 24.2 & 21.2 \\
        \rowcolor{Gray}
        T0++ (11B) & 35.4 & 33.3 \\
        \rowcolor{Gray}
        GPT-3  & 22.1 & 18.9 \\
        \rowcolor{Gray}
        GPT-3-NLI & 26.7 & 25.6 \\

        \rowcolor{Pink}
        CXR-BERT-NLI & 75.0	& 73.5 \\ 
       
        \rowcolor{Pink} SciFive$_{\text{large}}$-NLI & 47.5  & 35.4 \\
         \rowcolor{Pink} SciFive$_{\text{large}}$-NLI + SFT & 70.2  & 66.3 \\ 
        \rowcolor{Pink}  \modelsequential & 78.3  & 75.6 \\
        \textbf{\modellarge} &  \textbf{82.1} & \textbf{79.8} \\
        
        \midrule
        
        \multicolumn{3}{c}{\textbf{Biomedicine (MedNLI)}} \\
        \rowcolor{cambridgeblue}
        ESIM (MultiNLI) \cite{Chen2017EnhancedLF} & 51.7 & -\\

        \rowcolor{Gray}
        T0 (3B)  & 37.0 & 23.9 \\
        \rowcolor{Gray}
        T0++ (11B) & 55.2 & 44.6 \\
        \rowcolor{Gray}
        GPT-3  & 39.9 & 38.5 \\
        \rowcolor{Gray}
        GPT-3-NLI & 39.2 & 28.8\\
        \rowcolor{lightblue}

        \rowcolor{Pink}
         PubMedBERT-NLI & \textbf{75.7} &	\textbf{75.8} \\ 
        \rowcolor{Pink}\rowcolor{Pink} SciFive$_{\text{large}}$-NLI & 50.1 & 41.4 \\
        \rowcolor{Pink}
        SciFive$_{\text{large}}$-NLI  + SFT & 67.3 & 65.8 \\

        \rowcolor{Pink} \modelsequential & 71.4  & 71.5 \\
        \textbf{\modellarge} &  71.2 & 69.9\\
        \bottomrule
    \end{tabular}
    }
    \caption{Zero-shot \ac{NLI} results, showing micro accuracy and macro $F_1$. BERT and CXR-BERT are trained on MedNLI, we reproduce numbers from \citet{miura-etal-2021-improving} and \citet{boecking2022making} respectively. ESIM \cite{Chen2017EnhancedLF} is the highest-performing directly-transferred model reported by \citet{romanov2018lessons}. T0 \cite{sanh2022multitask} and GPT-3 \cite{brown2020language} baselines were conducted by us, matching stated hyperparameters where possible. Models with ``-NLI'' are fine-tuned or prompted baselines. `SFT' means with self-fine-tuning.}
    \label{table:zeroshot_nli}
\end{table}

\paragraph{Summarisation (\Cref{table:summarisation})} 

\modellarge\ achieves competitive performance compared with the best model in radiology (GPT-3-GW) and biomedical domains (T0 models) (\Cref{table:summarisation}).
In radiology, \modellarge\ is the second-best model. That the strongest performing models on summarisation are \acp{LLM} with substantially many parameters is not surprising; we observe in \Cref{sec:scale_up} that \model too enjoys scaling effects.
Most importantly, we again demonstrate the benefit from multi-task compositional transfer as \modellarge\ significantly outperforms both \modelsequential and SciFive-GW across all metrics in both domains. This further verifies that a na\"{i}ve sequential training on these two sources does not lead to effective compositional knowledge transfer.  
We also acknowledge it is more difficult to perform domain transfer for generation tasks in general: we cannot perform the data augmentation NLG and self-finetuning pipeline as it amounts to training the model to generate its own outputs.

\addtolength{\tabcolsep}{-2pt}
\begin{table}[tbh]
    \small
    \centering
    \scalebox{0.9}{
    \begin{tabular}{@{}lcccc@{}}
        \toprule[1.5pt]
        \textbf{Model} & \textbf{NEM} & \textbf{CheXbert} & \textbf{BLEU-4} & \textbf{ROUGE-L} \\
 
        \midrule[0.5pt]
        
        \multicolumn{5}{c}{\textbf{Radiology (Open-I Summarisation)}} \\
        \rowcolor{Gray}
        T0 (3B) & .054 & .243 & .027 & .088 \\
        \rowcolor{Gray}
        T0++ (11B)  & .019 & .145 & .012 & .061 \\
        \rowcolor{Gray}
        GPT-3 & .050 & .219 & .006 & .063 \\
        \rowcolor{Gray}
        GPT-3-GW & \textbf{.093} & \textbf{.304} & .019 & \textbf{.127} \\
        \rowcolor{Pink}
        SciFive$_{\text{large}}$-GW & .019 & .124 & .002 & .036  \\ 
        \rowcolor{Pink} \modelsequential & .050 & .256 & .015 & .077\\
        
        \addlinespace[5pt]
        \textbf{\modellarge} & .082 & .258 & \textbf{.038} & .117 \\
        
        \midrule[1pt]
        
        \multicolumn{5}{c}{\textbf{Biomedicine (PubMed ShortSum)}} \\
        \rowcolor{Gray}
        T0 (3B)                   & \textbf{.293} & - & .053 & .291 \\
        \rowcolor{Gray}
        T0++ (11B)                  & .290 & - & \textbf{.066} & \textbf{.341} \\
        \rowcolor{Gray}
        GPT-3                   & .197 & - &.017 & .184 \\
        \rowcolor{Gray}
        GPT-3-GW          & .272 & - &.046 & .266 \\
        \rowcolor{Pink}
        SciFive$_{\text{large}}$-GW  & .109 &  - & .010 & .149 \\
        \rowcolor{Pink}\modelsequential & .230 & - & .044 & .232\\
        \addlinespace[5pt]
        \textbf{\modellarge} & .263 & - &.047 & .260 \\
        
        \bottomrule[1.5pt]
    \end{tabular}
    }
    \caption{Zero-shot summarisation results. NEM (named entity matching) and CheXbert (radiology-specific) assess domain-specific factuality, while BLEU and ROUGE are standard lexical metrics. In all cases higher is better. GW = \textit{Gigaword}. T0~\cite{sanh2022multitask} and GPT-3~\cite{brown2020language} baselines were conducted by us.}
    \label{table:summarisation}
\end{table}
\addtolength{\tabcolsep}{2pt}

\paragraph{Text embedding learning (\Cref{table:text_embeddings})}

The \model-generated examples greatly improve the SOTA sentence embedding model's capability on both impression-to-findings retrieval in radiology and semantic textual similarity  (MedSTS) in the biomedicine domain (\Cref{table:text_embeddings}).
This is evidence that \model-generated sentences are of high quality and have captured semantic similarity and contradiction required for learning a good embedding model.  
We also compare with an ablated version of \model without in-domain MLM to generate data and find that the full model performs better across the board.
This shows the importance of domain training for generating good in-domain examples.
We explore this further in \Cref{sec:ablation}. 

\begin{table}[tbh]
    \small
    \centering
    \scalebox{0.83}{
    \begin{tabular}{@{}lccc@{}}
        \toprule[1.5pt]
        \multicolumn{4}{c}{\textbf{Radiology (Open-I Retrieval)}} \\
        \textbf{Model} & Acc$_{@1}$ & Acc$_{@5}$ & Acc$_{@10}$  \\
        \midrule[0.5pt]
        all-mpnet-base-v2 & 8.3 & 15.1 & 20.2 \\
        + \modellarge\  (no-MLM) & 12.0 & 19.9 & 22.8 \\ 
        + \modellarge  & \textbf{13.3} & \textbf{20.4} & \textbf{25.5} \\
        
        \midrule[1pt]
        
        \multicolumn{4}{c}{\textbf{Biomedicine (MedSTS)}} \\
        \textbf{Model} & $r$ & $\rho$ \\
        \midrule[0.5pt]
        all-mpnet-base-v2 & 72.8 & 64.6 \\
        + \model$_{\text{large}}$  (no-MLM) &  76.4$_{\pm0.04}$ & 67.1$_{\pm0.06}$  \\ 
        + \model$_{\text{large}}$  &  \textbf{76.9}$_{\pm0.00}$ & \textbf{67.9}$_{\pm0.09}$ \\
        \bottomrule[1.5pt]
    \end{tabular}
    }
    \caption{
        Text embedding learning results.
        Starting from a state-of-the-art embedding model (all-mpnet-base-v2), we fine-tune with \model-generated data (indicated by `+').
        Radiology evaluation is retrieval: given the \texttt{impression} section of a report, find the corresponding \texttt{findings} section.
        For biomedicine, we report similarity on MedSTS~\cite{yanshan2020medsts}, where $r$ and $\rho$ refer to Pearson's $r$ and $\rho$ (scaled by 100 for legibility).
    }
    \label{table:text_embeddings}
\end{table}

\section{Further Analysis}
\label{sec:analysis}

In this section, we demonstrate the importance of individual components of \model (\Cref{sec:ablation}) and explore the role of model size (\Cref{sec:scale_up}). Finally, we provide fine-grained analysis on RadNLI to verify whether \model has indeed acquired domain-specific task knowledge from compositional transfer (\Cref{sec:other_analysis}). 

\subsection{Ablation Study}\label{sec:ablation}
\begin{table}
    \small
    \centering
    \scalebox{0.83}{
    \begin{tabular}{lcc}
        \toprule
        Setting & RadNLI (acc.) & Sum. (NEM)  \\
        \midrule
        \modellarge\  (full model) &  \textbf{82.1} & \textbf{.082} \\
        \midrule
        (1) no in-domain MLM & 63.5 & .015 \\
        (2) no NLGU \& (3) & 59.0 & .052 \\
        (3) no self-finetuning & 49.6  & - \\
        \bottomrule
    \end{tabular}
    }
    \caption{
        Ablation study on \model components, evaluated on radiology. Removing \ac{MLM} removes in-domain text during pretraining. Removing NLGU reduces NLI to purely discriminative (thus also disabling self-finetuning) and summarisation to purely generative tasks. Self-finetuning is only used for NLI tasks. 
         Sum. = summarisation, NEM = named-entity matching metric. Note that the component is removed one by one but not incrementally.
    }
    \label{table:ablation_main}
\end{table}

Through ablations, we probe the contributions of key components of \model:
1) In-domain MLM,
2) NLGU (combining \ac{NLU} and \ac{NLG}) (\Cref{sec:in_dom_unsup_pretr}), and
3) self-finetuning for zero-shot \ac{NLI} (\Cref{sec:task_specific_inference}).
We conduct these ablations on the radiology domain on \modellarge.
The results are shown in \Cref{table:ablation_main}.

We observe that all components are essential to the success of the model.
In-domain \ac{MLM} is especially important for summarisation, without which the model fails in zero-shot transfer as it often just extracts a random subsequence of the document.
Removing NLGU harms both NLI and summarisation.
Training without NLGU removes the NLG component from NLI and therefore disables self-finetuning.
Self-finetuning is the most important component for boosting NLI performance, without which the model's accuracy drops more than 30\%. As shown in \Cref{table:zeroshot_nli}, SciFive also benefits from self-finetuning in this way.
This indicates that the pseudo in-domain NLI task data generated by NLGU is crucial.
Training without NLGU also removes the NLU task for summarisation and brings down the performance, indicating that having an NLU task can also benefit generation. 
 
\begin{figure}[t]
    \centering
    \includegraphics[width=0.8\linewidth]{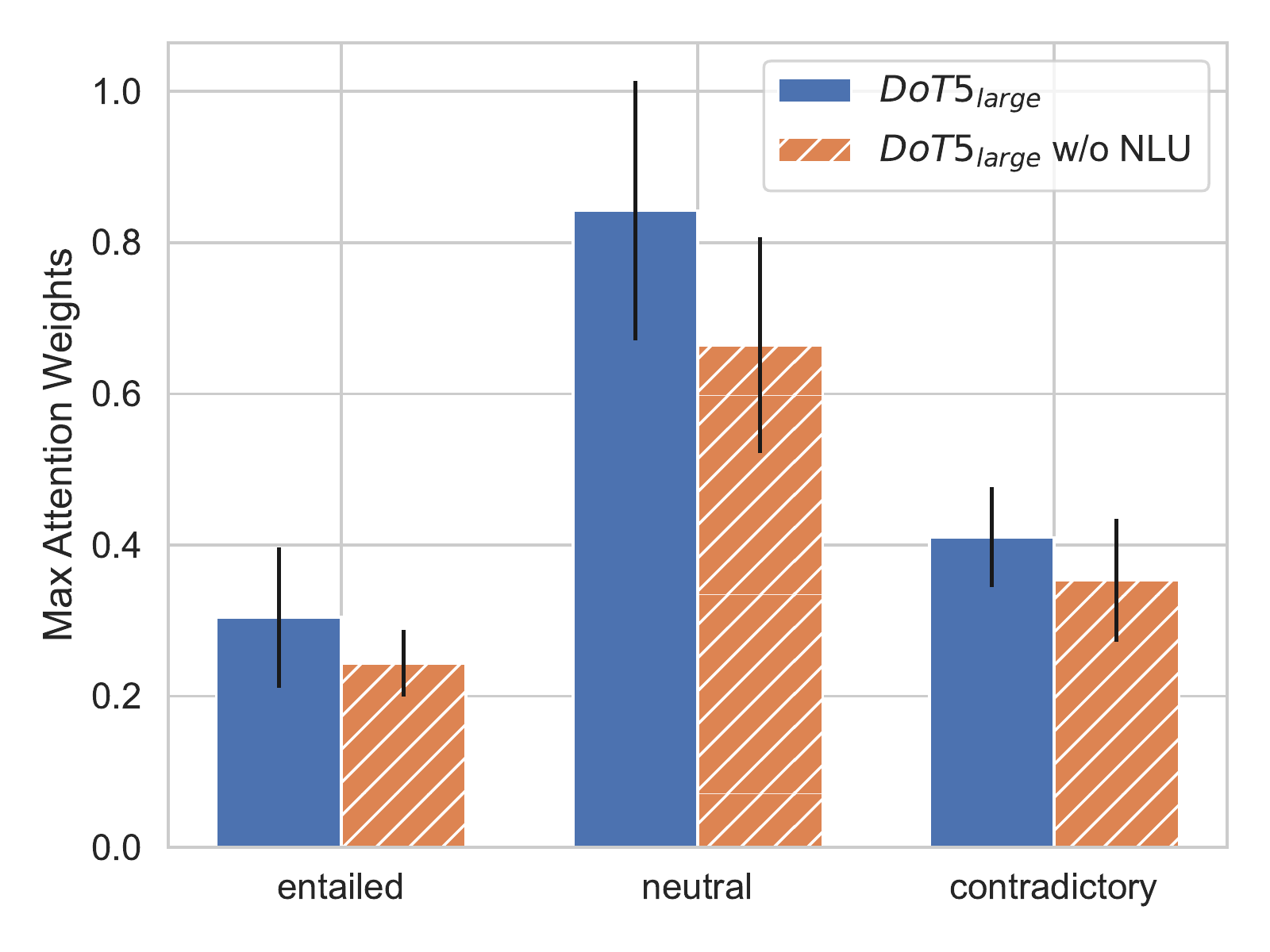}
    \caption{
        Maximum attention weights assigned to control code \{\texttt{label}\} (``entailed'', ``neutral'', ``contradictory'') in the prompt in NLI hypothesis generation, averaged over $100$ randomly sampled examples from the RadNLI dev set.
        Error bars represent standard deviation.
    }
    \label{fig:attention_nlgu}
\end{figure}

We hypothesise that NLU improves NLG by forcing the model to be more sensitive to the control code in the prompt, leading to improved pseudo-data generation and better summarisation.
To test this, following \citet{tang-etal-2018-analysis}, we compute the maximum attention weights across all attention heads to the control codes in the prompt when generating an NLI hypothesis (\Cref{fig:attention_nlgu}).
We compare \modellarge\ trained with or without NLU.
We see that the full model attends more on the control codes, suggesting that NLU is increasing label conditionality during generation.
\Cref{tab:nli_generated} shows some examples:
When required to generate an entailment, the model can usually correctly paraphrase the original sentence;
for negation, the model is usually able to correctly identify modifiers to flip the logic (e.g., change ``increase'' to ``decrease'' and adding or removing ``no'');
for neutral, the model generates a thematically related sentence but not directly negating or agreeing with the original sentence. 

\newcommand{\rot}[1]{\rotatebox{90}{#1}}

\begin{table}
    \scriptsize
    \centering
    \setlength{\tabcolsep}{4pt}
    \scalebox{0.95}{
    \centering
    \begin{tabular}{llll@{}}  
        \multicolumn{3}{@{}l}{\textbf{Prompt: Generate a} \{\texttt{label}\} \textbf{sentence of} \{\texttt{premise}\}:}\\
        \toprule
        \multicolumn{3}{@{}l}{{\texttt{premise}}: ``\textit{Low lung volumes but no evidence of pneumonia.}''} \\
        \addlinespace[2.5pt]
        
        \multirow{3}{*}{{\rot{\texttt{label}}}}
        & Entailment:     & ``\textit{There is no evidence of pneumonia.}''  \\
        & Neutral:        & ``\textit{The patient has a history of smoking.}'' \\ 
        & Contradiction:  & ``\textit{The patient has pneumonia.}'' \\ 
        
        \midrule
        \multicolumn{3}{@{}l}{{\texttt{premise}}: ``\textit{The patient is rotated slightly which limits assessment.}''} \\
        \addlinespace[2.5pt]
        \multirow{3}{*}{{\rot{\texttt{label}}}}
        & Entailment:     & ``\textit{Assessment is limited due to patient rotation.}''  \\
        & Neutral:        & ``\textit{The patient is rotated to the left.}'' \\ 
        & Contradiction:  & ``\textit{The patient is oriented perfectly. }'' \\
        
        \bottomrule
    \end{tabular}
    }
    \caption{
        Pseudo-\ac{NLI} data in the radiology domain generated by \modellarge\ for a given input premise and label.
        Premises are taken from the development split of the RadNLI dataset.
    }
    \label{tab:nli_generated}
\end{table} 

\subsection{Effect of Scaling Up}
\label{sec:scale_up}

\begin{figure}
    \centering
    \includegraphics[width=\linewidth]{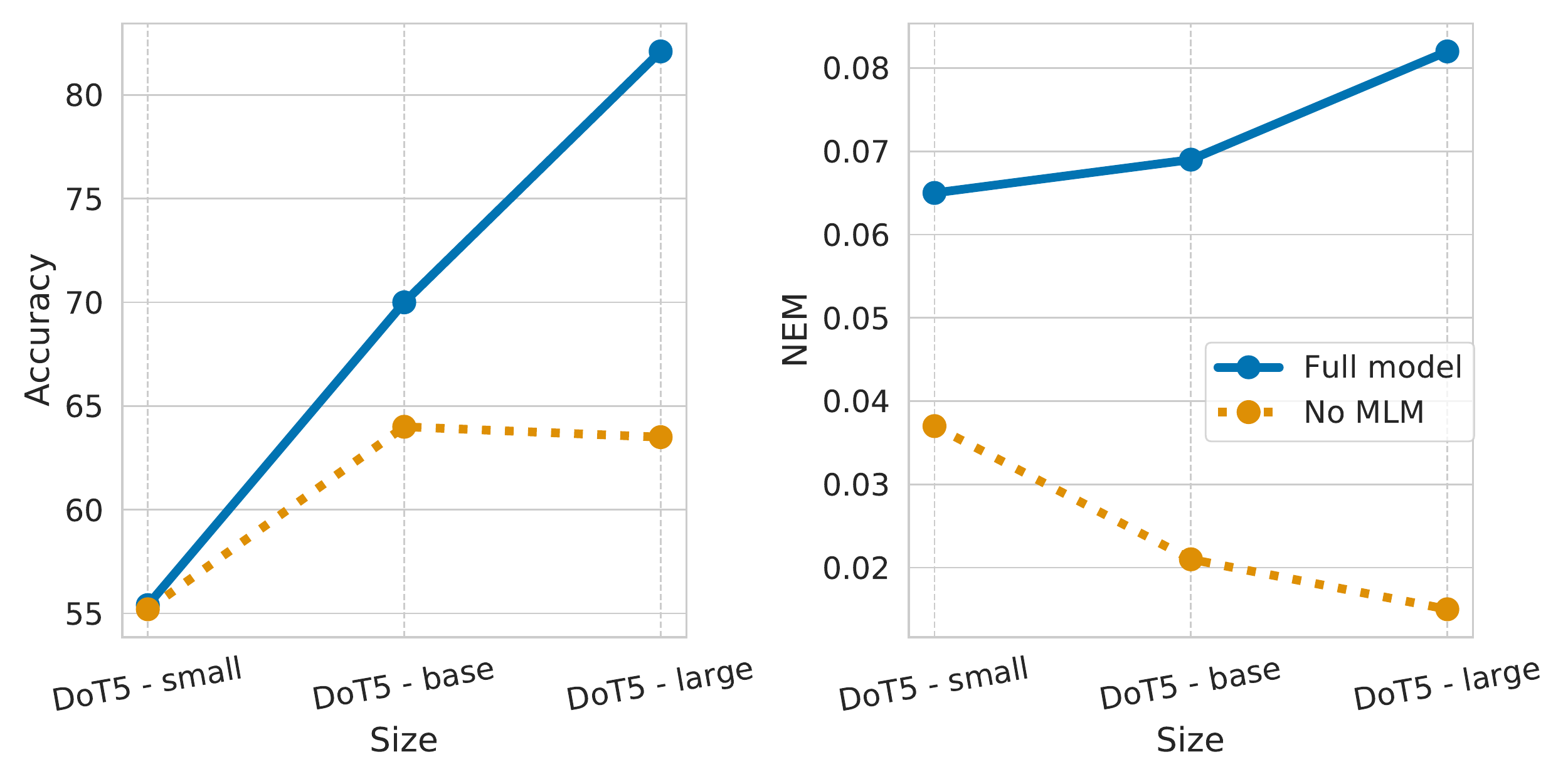}
    \caption{
        Scaling-up effect on RadNLI (left) and Open-I summarisation (right).
        Both the full model and its ablated versions are compared.
        Note that self-finetuning is only applicable for the \ac{NLI} tasks.
    }
    \label{fig:scale_up}
\end{figure}

We have so far reported results on a large T5 model (770M parameters).
In \Cref{fig:scale_up}, we plot the performance of small (70M) and base (220M) \model models with their ablated versions for RadNLI and radiology summarisation, showing a clear trend of increasing performance as the model size grows.
Interestingly, this scaling effect disappears when we remove in-domain MLM, revealing the importance of domain training for larger models, especially for summarisation.
This is possibly because, without domain training, scaling up the model leads to overfitting to the general-domain task data.
The compositional transfer framework from \model however regularises the model for more complex knowledge acquisition, and thus is able to harness the power from larger models.

\subsection{Evidence of Compositional Transfer in \model: A Case Study on RadNLI}
\label{sec:other_analysis}
Although RadNLI is a radiology-specific \ac{NLI} dataset, we observe that some examples may be solvable using general-domain task knowledge (e.g., syntactic cues) alone.
A general-purpose \ac{NLI} model will likely detect that `There is no pneumothorax' contradicts `There is a pneumothorax' without requiring radiology-specific knowledge such as an understanding of pneumothorax.
Therefore, higher performance on RadNLI may not strictly guarantee the model has acquired in-domain knowledge.
To quantify how much of \model's transfer success is due to the acquisition of the previously unseen domain-specific task knowledge versus from direct application of the general-domain task knowledge, we manually annotated each of the 480 sentence pairs in the RadNLI test set by whether it could be solved without particular medical expertise%
\footnote{We determined 228 (47\%) pairs could be solved without medical/radiological expertise, 177 (37\%) could not, and the remaining 75 (16\%) were ambiguous. Ambiguous cases were excluded from the analysis.
}.
Examples are shown in \Cref{table:radnli_examples}.

\begin{table}[t]
\small
    \centering
    \scalebox{0.9}{
    \begin{tabular}{@{}ll@{}}
    \toprule
    \multicolumn{2}{@{}l@{}}{\textbf{Does not require radiology expertise}}\\
        Premise &  \textit{There is a small left pleural effusion.}\\
        Hypothesis & \textit{No pleural effusion or pneumothorax is seen.}\\
        Label & Contradiction \\
        \midrule
        \multicolumn{2}{@{}l@{}}{\textbf{Requires radiology expertise}}\\
        Premise & \textit{The cardiac silhouette is top normal.} \\
        Hypothesis & \textit{The heart is not enlarged.} \\
        Label & Entailment \\
        \bottomrule
    \end{tabular}
    }
    \caption{
        Examples from RadNLI that do or do not require radiology-specific knowledge to solve.
        While all models listed in \Cref{tab:radnli_analysis} correctly solved the top example, only \modellarge\ solved the more challenging second example.
    }
    \label{table:radnli_examples}
\end{table}

\begin{table}
    \small
    \centering
    \scalebox{0.9}{
    \begin{tabular}{@{}l c| c c@{}}
    \toprule
        &  & \multicolumn{2}{c}{\textbf{Expertise required}} \\
        \textbf{Model} &  \textbf{All cases} & \textbf{Yes} & \textbf{No} \\
        \midrule
        a) \modellarge & \textbf{80.7} & \textbf{70.1} & \textbf{86.4} \\
        \hspace{5pt} No self-finetuning & 51.0 & 43.3 & 50.8 \\
        \midrule
        b) \modelsequential & 75.6  & 54.8 & \textbf{86.5} \\
        \hspace{5pt} No self-finetuning  & 35.6 & 36.2 & 35.6 \\
        \midrule
        c) T5$_\text{large}\rightarrow$Task & 59.5 & 35.3 & 70.1 \\
        \hspace{5pt} No self-finetuning & 37.5 & 36.1 & 35.1 \\
        \midrule
        Zero-rule baseline & 24.6 & 29.0 & 20.0 \\
        \bottomrule
    \end{tabular}
    }
    \caption{
        Macro $F_1$ of \model with and without in-domain data during pretraining, on subsets of RadNLI requiring radiology-specific expertise or not.
        The zero-rule baseline always outputs the most common class (for RadNLI, this is `Neither').
        We report macro $F_1$ to account for differing label distributions. Note that T$5_\text{large}\rightarrow$Task is equivalent to \modellarge without in-domain MLM training.}
    \label{tab:radnli_analysis}
\end{table}

\Cref{tab:radnli_analysis} compares three models on these subsets: \modellarge (a), \modelsequential (b), and \modellarge\ \emph{without} in-domain MLM (c) (equivalent to `T5$_\text{large}\rightarrow$Task'). We further test with and without self-finetuning to probe its capacity to strengthen domain-specific competence.

While \modellarge\ achieves the best performance overall, it is specifically on challenging domain-specific cases that it outperforms \modelsequential, an increase of 15 points in $F_1$.
For example, in \Cref{table:radnli_examples}, only \modellarge\ is able to solve the second example which requires radiology-specific knowledge (the model should know cardiac silhouette includes heart size; and if the heart is top normal, then it should not be enlarged).
This demonstrates the role of compositional transfer for inferring the otherwise unseen in-domain task knowledge (in this case, radiology \ac{NLI} knowledge) solving challenging cases that require expertise.

The two ablated versions help understand where this domain-specific task knowledge is acquired.
In-domain \ac{MLM} training is key as removing it (c) significantly decreases the performance on domain-expert cases in particular, producing a model which cannot benefit from self-finetuning at all for such cases. This is because without in-domain MLM, the model is not able to generate good-quality pseudo in-domain labels in the first place, and therefore self-finetuning has little effect on the expert cases. Introducing in-domain data sequentially (b) resolves the performance gap on non-expert cases, but still underperforms on domain-specific cases relative to multi-task training (a).
We conclude that the compositional fusion of task and domain knowledge happens during \model's multi-task pretraining phase with in-domain MLM as the key element, and that domain-specific competence is elicited through self-finetuning. 
    \section{Conclusion and Discussion}
\label{sec:conc}

We propose \model, a compositional transfer learning framework to solve domain-specific NLP tasks without requiring in-domain task labels.
We show the effectiveness of \model on zero-shot transfer to multiple tasks in the biomedicine and radiology domains.
\model significantly outperforms T5 sequential training across all tasks, and achieves zero-shot \ac{SOTA} in radiology NLI with massive gains. 
We also conduct extensive analyses to identify the contribution from each model component and the benefits from scaling up the model size, and demonstrate direct evidence of domain-specific task knowledge learned from \model's compositional transfer.

Limitations of this work include the challenge of drawing clear boundaries between domains and the necessarily incomplete exploration of hyperparameters and configurations. For example, general domain texts may contain biomedical or radiology sources, and our `biomedical' \ac{NLI} evaluation set leans strongly clinical, introducing a degree of domain shift. Investigation of the weighting of terms in the loss reveals the potential to improve performance through more exhaustive hyperparameter search - we emphasise that this was a proof-of-concept study and although \model performs favourably, zero-shot domain transfer could be further pushed, especially if only a single downstream task is required.

The proposed NLGU method and subsequent self-finetuning was critical for improving downstream task performance. However, we observed an intermittent negative effect wherein the model would attempt to solve the \ac{NLU} task when presented with an unusually long prompt. Further work can be done to refine this approach. For example, the benefit of NLGU in resource-rich domains is unclear. As our focus is on domain transfer and we do not evaluate on general-domain tasks, we leave such experimentation to future study.

Finally, we acknowledge that it is non-trivial to apply our full framework to single-sentence/paragraph classification tasks. While our most basic setup (compositional training of in-domain MLM and vanilla task training) can still be transferable to any task format, NLGU and self-finetuning would currently only work for tasks that involve pairs of texts.
Nonetheless, we believe \model proves to be a highly effective zero-shot domain transfer framework which will be beneficial to domain-specific applications beyond radiology and biomedicine.

\section*{Acknowledgments}
The authors would like to thank the anonymous TACL reviewers and editors for their detailed feedback and helpful suggestions.
    
    \balance
    \bibliography{main}
    \bibliographystyle{acl_natbib}
    
\end{document}